\documentclass[10pt,twocolumn,letterpaper]{article}

\usepackage{iccv}
\usepackage{times}
\usepackage{epsfig}
\usepackage{graphicx}
\usepackage{amsmath}
\usepackage{amssymb}
\usepackage{multirow}
\usepackage{makecell}
\usepackage{balance}
\usepackage{rotating}
\usepackage{booktabs}

\usepackage[pagebackref=true,breaklinks=true,letterpaper=true,colorlinks,bookmarks=false]{hyperref}

\iccvfinalcopy 

\def\iccvPaperID{6948} 

\ificcvfinal\fi

\newcommand{\midsepremove}{\aboverulesep = 0mm \belowrulesep = 0mm}
\midsepremove

\begin{document}

\title{Knowledge-Enhanced Hierarchical Information Correlation Learning \\ for Multi-Modal Rumor Detection}

\author{\textbf{Jiawei Liu}, \textbf{Jingyi Xie}, \textbf{Fanrui Zhang}, \textbf{Qiang Zhang}, \textbf{Zheng-jun Zha}\\[\bigskipamount]
University of Science and Technology of China\\
{jwliu6@ustc.edu.cn, \{hsfzxjy, zfr888, zq\_126\}@mail.ustc.edu.cn, zhazj@ustc.edu.cn}
}

\maketitle
\ificcvfinal\thispagestyle{empty}\fi

\begin{abstract}
   The explosive growth of rumors with text and images on social media platforms has drawn great attention. Existing studies have made significant contributions to cross-modal information interaction and fusion, but they fail to fully explore hierarchical and complex semantic correlation across different modality content, severely limiting their performance on detecting multi-modal rumor. In this work, we propose a novel knowledge-enhanced hierarchical information correlation learning approach (KhiCL) for multi-modal rumor detection by jointly modeling the basic semantic correlation and high-order knowledge-enhanced entity correlation. Specifically, KhiCL exploits cross-modal joint dictionary to transfer the heterogeneous unimodality features into the common feature space and captures the basic cross-modal semantic consistency and inconsistency by a cross-modal fusion layer. Moreover, considering the description of multi-modal content is narrated around entities, KhiCL extracts visual and textual entities from images and text, and designs a knowledge relevance reasoning strategy to find the shortest semantic relevant path between each pair of entities in external knowledge graph, and absorbs all complementary contextual knowledge of other connected entities in this path for learning knowledge-enhanced entity representations. Furthermore, KhiCL utilizes a signed attention mechanism to model the knowledge-enhanced entity consistency and inconsistency of intra-modality and inter-modality entity pairs by measuring their corresponding semantic relevant distance. Extensive experiments have demonstrated the effectiveness of the proposed method.
\end{abstract}

\section{Introduction}
   
   \begin{figure}[t]
       \centering
       \includegraphics[width=\linewidth]{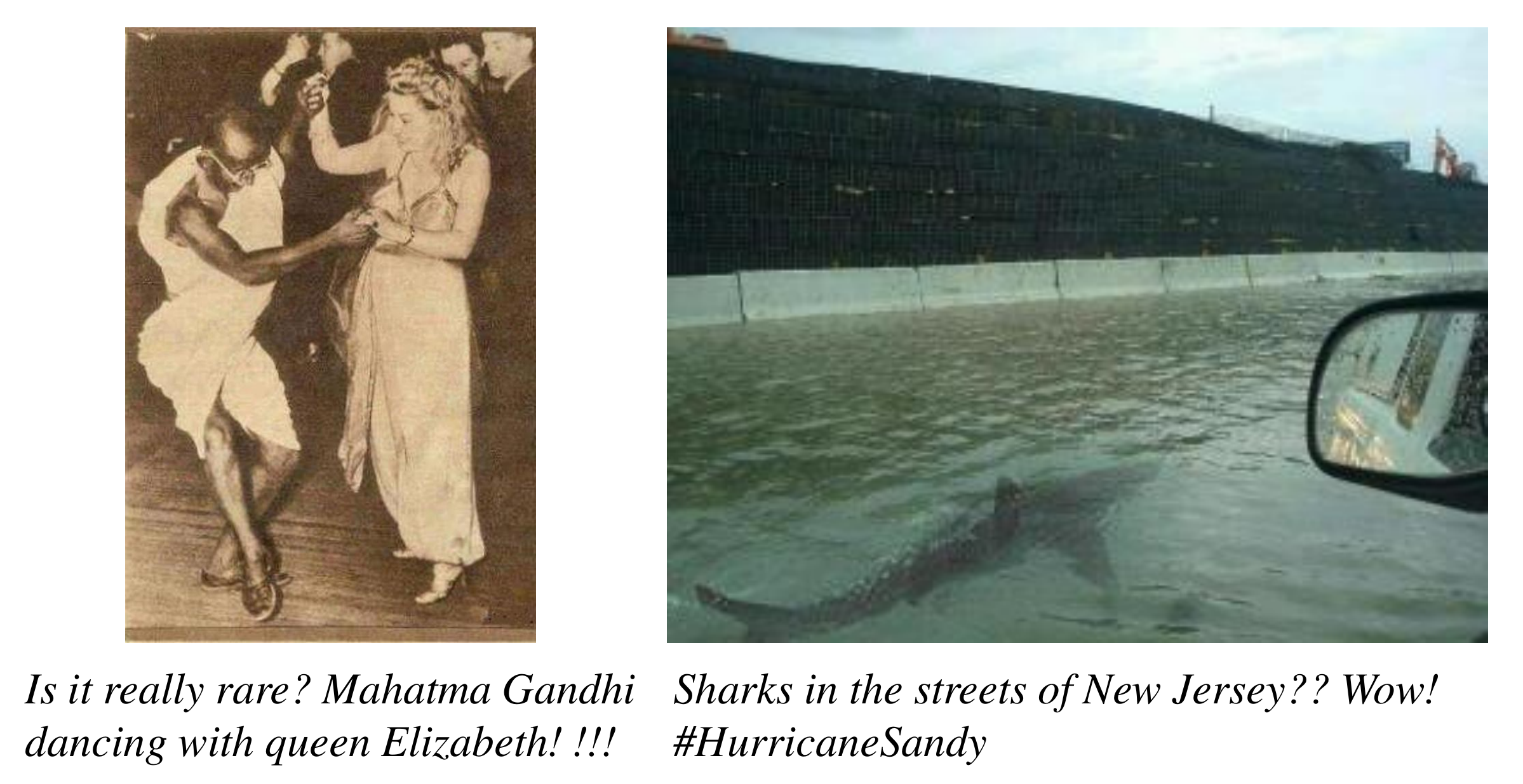}
       \caption{A real-world example of two fake multimedia tweets. The valuable high-order entity correlations in the two tweets provide diverse pivotal clues for detection.}
       \label{fig1}
   \end{figure}
   
   
   \label{sec:intro}
   The emergence of social media has revolutionized the conventional way people acquire information online. People enjoy the convenience and efficiency of online social media in sharing information and exchanging viewpoints \cite{abdelnabi2022open,mishra2020fake,huh2018fighting}. Unfortunately, it has also become much easier for fostering varied false rumors, which often include misrepresented or forged multimedia content. The widespread social media rumors can cause massive panic and
   social unrest, \textit{e.g.}, some lawbreakers utilize rumors to mislead health-protective behaviors and damage the government's credibility \cite{narayan2022desi,zhou2021joint,agarwal2019protecting}. Therefore, it is desirable to automatically detect and regulate rumors to promote truthful information in the social media platforms.
   
   
   Traditional rumor detection approaches typically concentrate on text-only content analysis, \textit{e.g.,} the news headline and body, user interactions over social media. These methods can be summarized into two categories, 1) the first one designs complicated hand-crafted features from the content of posts or news \cite{castillo2011information,feng2012syntactic,perez2017automatic}, 2) and the second one extracts deep features from deep neural networks (DNN) based models \cite{liu2018early,khoo2020interpretable}. As multimedia technology develops, rumor spreaders tend to utilize visual content together with textual content to draw more attention and obtain rapid diffusion. Multi-modal rumor detection has become an newly emerging and attractive task.  Compared to the single-modal rumor detection, it is more challenging to learn effective feature representation from heterogeneous multi-modal data, while also provides abundant complementary clues to facilitate detection false rumors. This work thus targets multi-modal rumor detection, which leverages multiple modalities of text and images to verify rumors.
   
   To address this task, a line of multi-modal rumor detectors have been proposed to explore cross-modal information interaction and fusion to identify anomalies in rumors. Most of them only model the basic semantic correlation between images and text at the feature level, ignoring high-order entity correlation across different modality content. These approaches simply utilize concatenation operation \cite{singhal2019spotfake,wang2018eann,wang2021multimodal}, attention mechanism \cite{jin2017multimodal,zhou2022multimodal}, or auxiliary tasks \cite{khattar2019mvae,zhou2020safe,chen2022cross} to capture basic semantic correlation between visual and textual features and generate multi-modal representation. In fact, multimedia content is narrated around entities, the knowledge-level correlations of entities are also essential for predicting the veracity of a sample, apart from the basic semantic correlation at the feature level \cite{dun2021kan}. To assist model high-order correlations among entities, it is indispensable to incorporate the complementary background knowledge information contained in external knowledge graph. A knowledge graph (KG) is composed of entities as graph nodes and relationships as edges with different types. For example, in Figure \ref{fig1},  it is beneficial to judge the falseness of multimedia posts by considering the knowledge-level entity inconsistency in cross-modality that the textual entity \textit{Elizabeth} and the visual entity \textit{woman} in the image are mismatched for the first post, and the knowledge-level entity inconsistency in intra-modality that the entities \textit{sharks} and \textit{street} rarely co-occur in real world for the second post, given the background semantic relevance in knowledge graph. Such valuable background semantic relevance in knowledge graph can't be directly learned from multimedia posts. 
   
   Although a few of methods attempt to introduce knowledge graph into rumor detection model, they aims at enriching individual entity by incorporating the knowledge information from all its one-hop neighborhood nodes \cite{dun2021kan,tseng2022kahan,wang2020fake}, or roughly consider the cross-modal relationship between visual and textual entities by calculating pairwise embedding distance \cite{qi2021improving,cui2020deterrent,li2021entity}. Therefore, these methods either completely ignore the knowledge-level correlation of entities or can not capture high-order correlation of intra and inter-modality entity pairs according to their different background semantic relevance in knowledge graph.
   
   In this work, we propose a knowledge-enhanced hierarchical information correlation learning approach (KhiCL) for multi-modal rumor detection to judge the authenticity of multimedia posts. It jointly models the basic semantic correlation and high-order knowledge-enhanced entity correlation to learn effective semantic-level and knowledge-level rumor representations. Specifically, KhiCL consists of a Basic Semantic Correlation Module (BSC), a Knowledge-Enhanced Entity Correlation Module (KEC) and a backbone network. BSC firstly exploits cross-modal joint dictionary II a transformer layer, in which the learnable set of modality-shared atoms makes the heterogeneous features consistent in a common feature space. BSC then utilizes a cross-modal fusion layer composed of \textit{Compare} and \textit{Aggregate} operations to perform cross-modal information interaction and fusion for further learning the basic cross-modal semantic inconsistency and consistency, respectively. Moreover, KEC identifies visual and textual entities in the multimedia posts, and designs a knowledge relevance reasoning strategy to find the shortest semantic relevant path between each pair of entities in external knowledge graph, and absorbs all complementary contextual knowledge of the connected entities in this path to learn knowledge-enhanced entity representations. It measures the semantic relevant distance for each pair of entities and selects top-$k$ consistency and inconsistency entity pairs. These selected entity pairs are further fused by utilizing the signed attention mechanism to capture both positive (consistency) and negative (inconsistency) knowledge-enhanced entity correlations, simultaneously. Extensive experiments on two benchmarks demonstrate our method outperforms the state-of-the-art methods by a large margin.
   
   The main contributions of this paper are as following: (1) We propose a knowledge-enhanced hierarchical information correlation learning approach for multi-modal rumor detection by jointly modeling the ba sic semantic correlation and high-order knowledge-enhanced entity correlation. (2) We propose a Basic Semantic Correlation Module with a cross-modal joint dictionary and a cross-modal fusion layer to reduce feature granularity gap and model the basic semantic correlation. (3) We design a Knowledge-Enhanced Entity Correlation Module with a knowledge relevance reasoning strategy and a signed attention mechanism to capture high-order knowledge-enhanced entity correlation.
   
   \begin{figure*}[t]
       \centering
       \includegraphics[width=\linewidth]{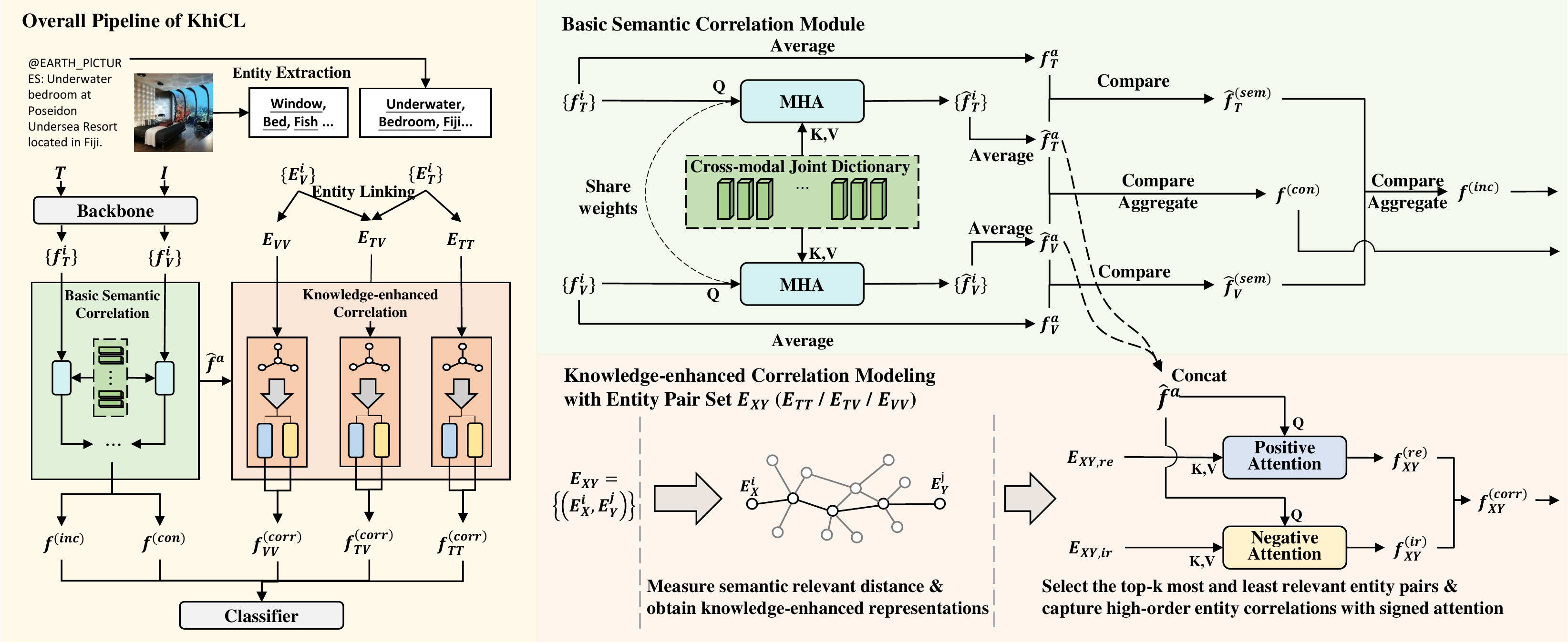}
       \caption{The overall architecture of the proposed KhiCL. It contains three components: a Basic Semantic Correlation Module (BSC), a Knowledge-Enhanced Entity Correlation Module (KEC) and a backbone network.}
       \label{fig:overoframework}
   \end{figure*}
   
   \section{Related Work}
   
   \textbf{Rumor Detection on Text.} Rumor detection has become an active research topic. Researchers regard it as a kind of binary classification task, and propose various methods using text-only modality. Early works mainly focus on designing sophisticated hand-crafted features from text content \cite{castillo2011information,feng2012syntactic} or user interactions based on social networks \cite{wu2015false,ma2015detect}, such as lexical and syntactic features \cite{perez2017automatic}. Recent works \cite{liu2018early} that employed deep learning models achieve promising results on detecting rumor with texts. For example, Khoo \textit{et al.} \cite{khoo2020interpretable} proposed a post-level attention framework (PLAN) to model long distance interactions between tweets by utilizing multi-head attention mechanism.
   
   
   \textbf{Multi-Modal Rumor Detection.} Recently, multi-modal rumor detection has received considerable attentions since the content form of posts tends to coexist with multi-modal information. For example, Wang \textit{et al.} \cite{wang2018eann} proposed to concatenate the visual and textual representations of post for obtaining a multi-modal features and identifying rumors. However, these earlier methods \cite{singhal2019spotfake} adopt simple fusion strategy and can not effectively capture semantic correlation between visual and textual features. 
   
   Considering that, several methods design sophisticated attention mechanism \cite{wu2021multimodal,qian2021hierarchical, zhou2022multimodal} or leverage auxiliary tasks \cite{khattar2019mvae,zhou2020safe,chen2022cross} to further explore semantic correlation between two modality features. For example, Wu \textit{et al.} \cite{wu2021multimodal} stacked multiple co-attention layers to fuse the multi-modal features, which could explore inter-dependencies among them. 
   Khattar \textit{et al.} \cite{khattar2019mvae} proposed an Multimodal Variational Autoencoder, which introduces a Variational Autoencoder to reconstruct visual data according to the input multi-modal information and learns the multi-modal representation. Zhang \textit{et al.} \cite{zhang2021multi} proposed a multi-modal meta multi-task learning method, which fully uses the stance information of user comments and learns a precise multi-modal representations of posts. Nevertheless, these attention mechanism and auxiliary task based methods only focus on basic semantic correlation at the feature level, they ignore high-order entity correlation across different modality content.
   
   
   \textbf{Knowledge Graph.} Knowledge graph is a structured data model, which has millions of entries that describe real-world entities, \textit{e.g.}, \textit{people}, \textit{products} and \textit{locations}. Entities in knowledge graph are represented as graph nodes, and the relations between entities are represented as edges. Knowledge graph has been utilized in many practical applications, such as goods recommendation \cite{zhang2016collaborative}, and question answering \cite{su2018learning,narasimhan2018straight}. A few of methods attempt to introduce knowledge graph into rumor detection to supplement the semantic integrity of post contents to obtain better representation \cite{dun2021kan,tseng2022kahan,qi2021improving,cui2020deterrent,li2021entity}. For example, Dun \textit{et al.} \cite{dun2021kan} proposed Knowledge-aware Attention Network that incorporates entities and their entity neighbour contexts to provide complementary information. Qi \textit{et al.} \cite{qi2021improving} explicitly extracted the visual entities to explore the high-level semantics of images and models the multi-modal inconsistency and enhancement by calculating pairwise similarity of multi-modal entities. These above-mentioned methods either neglect the knowledge-level correlation of entities or can not capture high-order semantic correlation of intra and inter-modality entitiy pairs according to the specific background semantic relevance in knowledge graph.

   \section{Method}
   
   
   
   Multi-modal rumor detection is usually formulated as a binary classification problem to judge the given posts with texts and images as rumor or non-rumor. To address the problem, we propose a knowledge-enhanced hierarchical information correlation learning approach to jointly model the basic semantic correlation and high-order knowledge-enhanced entity correlation. As illustrated in Figure \ref{fig:overoframework}, it consists of a BSC and a KEC, followed after a backbone network. Given a post with image $\boldsymbol{\boldsymbol{I}}$ and text $\boldsymbol{\boldsymbol{T}}$, the backbone network is firstly adopted to encode the  basic visual and textual features. The two features are then fed into BSC, which employs a transformer with cross-modal joint dictionary to reduce the modality gap and reconstruct the visual and textual representations. BSC further utilizes a cross-modal fusion layer to estimate basic cross-modal semantic inconsistency and consistency information. Moreover, KEC identifies visual and textual entities in the multimedia posts and leverages a knowledge relevance reasoning strategy to learn knowledge-enhanced entity representations. It then measures the semantic relevant distance for each pair of entities and filters top-$k$ consistency and inconsistency entity pairs, which are further fused by the signed attention mechanism to capture both consistent and inconsistent knowledge-enhanced entity correlations. The basic cross-modal semantic correlations along with knowledge-enhanced entity correlations are then consumed by a classifier, which jointly judges the authenticity of a post.
   
   
   \subsection{The Backbone Network}
   
   The backbone network is designed as a two-stream fashion, consisting of a textual encoder and a visual encoder to extract basic features from the text $\boldsymbol{\boldsymbol{T}}$ and image $\boldsymbol{\boldsymbol{I}}$.
   
   \textbf{Textual Encoder.} Specifically, we employ a pretrained BERT model \cite{devlin2018bert} to map the word sequence of $\boldsymbol{\boldsymbol{T}}$ into an embedding sequence $\{\boldsymbol{w}_t^i\}_{i=1}^L$ of length $L$. The embedding sequence are consumed by a bidirectional long short term memory network (Bi-LSTM) \cite{litman2020scatter}, which is further transformed by a linear layer to obtain the textual feature sequence $\{\boldsymbol{f}^i_T\}_{i=1}^L \in \mathbb{R}^d$. It is formulated as following: 
   \begin{align}
     \{\boldsymbol{f}^i_T\} = \boldsymbol{W}_t^T (\text{Bi-LSTM}({\boldsymbol{w}_t^i}))+\boldsymbol{b}_t
   \end{align}
   where $\boldsymbol{W}_t$ and $\boldsymbol{b}_t$ denote the learnable parameters of a fully connected (FC) layer. 
   
   \textbf{Visual Encoder.} ResNet-50 model \cite{he2016deep} is utilized to encode image $\boldsymbol{\boldsymbol{I}}$ into a initial feature map with dimension $2048 \times 7 \times 7$, which is then transformed by a convolutional block and reshape operation into the visual feature sequence $\{\boldsymbol{f}^i_V\}_{i=1}^{49} \in \mathbb{R}^d$. This is formulated as following: 
   \begin{equation}
     \begin{aligned}
       \{\boldsymbol{f}^i_V\}  = \text{Reshape}(\text{ReLU}(\text{BN}(\text{Conv}(\text{CNN}(\boldsymbol{I})))))
     \end{aligned}
   \end{equation}
   where $\text{CNN}$ denotes ResNet-50 model, $\text{Conv}$ represents a convolutional layer. $\text{BN}$ represents a batch normalization layer, and $\text{ReLU}$ refers to a rectified linear unit layer. The obtained textual and visual feature sequences $\{\boldsymbol{f}^i_T\}_{i=1}^L$ and $\{\boldsymbol{f}^i_V\}_{i=1}^{49}$ are regarded as input to BSC.
   
   \subsection{Basic Semantic Correlation}
   Basic Semantic Correlation Module is responsible to model the basic semantic correlation from the visual and textual feature sequences  $\{\boldsymbol{f}^i_T\}$ and $\{\boldsymbol{f}^i_V\}$. Features from images and texts exist large semantic gap, so that it is significant to align the different modality features by transforming the heterogeneous features into a shared space, prepared for capturing more informative cross-modal correlation information. BSC firstly shrinks the modality gap between the two sequences by fundamentally introducing a cross-modal joint dictionary containing a learnable set of modality-shared atoms, which combines with a transformer layer to reconstruct the representations. It ensures the heterogeneous features to be consistent in a common feature space. Afterwards, BSC further utilizes a cross-modal fusion layer to perform cross-modal information interaction and fusion between original and reconstructed representations to learn the basic cross-modal semantic inconsistency and consistency.
   
   \textbf{Cross-Modal Alignment.} Specifically, a cross-modal joint dictionary $\boldsymbol{D}$ consists of $M$ learnable modality-shared atoms $\{\boldsymbol{D}_i\}_{i=1}^M \in \mathbb{R}^d$, each of which represents a modality-agnostic semantic pattern. A shared multi-headed attention transformer layer (MHA) is utilized to reconstruct the representations for both modalities in a common feature space, which consists of a multi-head self-attention module and a fully connected feed-forward network. The transformer layer takes respective feature sequence $\{\boldsymbol{f}^i_T\}$ or $\{\boldsymbol{f}^i_V\}$ as query, and the cross-modal joint dictionary $\boldsymbol{D}$ as key and value to learn the corresponding semantically aligned feature sequence $\{\hat{\boldsymbol{f}}_T^i\}$ or $\{\hat{\boldsymbol{f}}_T^i\}$. This is formulated as follows:
   \begin{equation}
     \begin{aligned}
       \{\hat{\boldsymbol{f}}_T^i\} & =\text{MHA}(\{{\boldsymbol{f}}_T^i\}, \{\boldsymbol{D}_i\}_{i=1}^M, \{\boldsymbol{D}_i\}_{i=1}^M) \\
       \{\hat{\boldsymbol{f}}_V^i\} & =\text{MHA}(\{{\boldsymbol{f}}_V^i\}, \{\boldsymbol{D}_i\}_{i=1}^M, \{\boldsymbol{D}_i\}_{i=1}^M)
     \end{aligned}
   \end{equation}
   
   \textbf{Modeling Basic Semantic Correlation.} In order to aggregate global semantic information over corresponding sequence, in the cross-modal fusion layer, the features $\{\boldsymbol{f}^i_T\}, \{\boldsymbol{f}^i_V\}, \{\hat{\boldsymbol{f}}_T^i\}$ and $\{\hat{\boldsymbol{f}}_V^i\}$ are sequentially and spatially collapsed into $d$-dimensional vectors. We apply a average pooling layer and a FC layer to abstract the aggregated features, which is calculated as following:
   \begin{equation}
   \begin{aligned}
    {\boldsymbol{f}}_T^a, \;  \hat{\boldsymbol{f}}_T^a &=\boldsymbol{W}_a^T(\text{Avg}(\{\boldsymbol{f}_T\}, \{\hat{\boldsymbol{f}}_T\}), ) + \boldsymbol{b}_a \\
     {\boldsymbol{f}}_V^a, \; \hat{\boldsymbol{f}}_V^a &=\boldsymbol{W}_a^T(\text{Avg}(\{\boldsymbol{f}_V\}, \{\hat{\boldsymbol{f}}_V\}), ) + \boldsymbol{b}_a
   \end{aligned}
   \end{equation}
   where $\boldsymbol{W}_a$ and $\boldsymbol{b}_a$ are the parameters of the FC layer. The cross-modal fusion layer learns semantic inconsistency representation $\boldsymbol{f}^{(inc)}$ and consistency representation  $\boldsymbol{f}^{(con)}$ from the obtained aggregated features $\boldsymbol{f}_T^a, \boldsymbol{f}_V^a, \hat{\boldsymbol{f}}_T^a$ and $\hat{\boldsymbol{f}}_V^a$ by employing \textit{Compare} and \textit{Aggregate} operations, which excavate interaction and fusion among these features. The representation $\boldsymbol{f}^{(inc)}$ measures the inter-modal inconsistency of the two modalities. Concretely, the cross-modal fusion layer firstly learns the modality-aware semantic features $\boldsymbol{f}_T^{(sem)}$ and $\boldsymbol{f}_V^{(sem)}$ by calculating the discrepancy between the original feature with the reconstructed feature in respective modality,
   \begin{align}
     \boldsymbol{f}_T^{(sem)}=\boldsymbol{f}_T^a-\hat{\boldsymbol{f}}_T^a, \;
     \boldsymbol{f}_V^{(sem)}=\boldsymbol{f}_V^a-\hat{\boldsymbol{f}}_V^a
   \end{align}
   The features $\boldsymbol{f}_T^{(sem)}$ and $\boldsymbol{f}_V^{(sem)}$ own semantic information that is unique to the corresponding modality, which are compared to form the semantic inconsistency representation $\boldsymbol{f}^{(inc)}$ as following:
   \begin{align}
     \boldsymbol{f}^{(inc)} = [\boldsymbol{f}_T^{(sem)}; \boldsymbol{f}_T^{(sem)}-\boldsymbol{f}_V^{(sem)}; \boldsymbol{f}_V^{(sem)}]
   \end{align}
   Similarly, this layer gathers and concatenates the modal-agnostic features $\hat{\boldsymbol{f}}_T^a$ and $\hat{\boldsymbol{f}}_V^a$ to composite the representation $\boldsymbol{f}^{(con)}$ as following:
   \begin{align}
     \boldsymbol{f}^{(con)} = [\hat{\boldsymbol{f}}_T^a; \hat{\boldsymbol{f}}_T^a \odot \hat{\boldsymbol{f}}_V^a; \hat{\boldsymbol{f}}_V^a]
   \end{align}
   The representation $\boldsymbol{f}^{(con)}$ captures the joint semantic consistency for the post. Finally, the representations $\boldsymbol{f}^{(inc)}$ and $\boldsymbol{f}^{(con)}$, carrying basic cross-modal semantic inconsistency and consistency, are adopted as part of the rumor representation to judge the authenticity of the post. Moreover, the modal-agnostic features $\hat{\boldsymbol{f}}_T^a$ and $\hat{\boldsymbol{f}}_V^a$ are further leveraged by KEC for knowledge-level correlation modeling.
   
   \subsection{Knowledge-Enhanced Entity Correlation}
   Multi-modal entity correlation is a pivotal indicator for rumors. KEC distills background knowledge information from real-world knowledge graph to complement the semantic correlation, and captures high-order knowledge-enhanced entity correlation for multimedia posts. It firstly identifies visual and textual entities from images and texts, which are then linked to generate one inter-modality entity pair set (visual-textual) and two intra-modality entity pair sets  (visual-visual and textual-textual). A novel knowledge relevance reasoning strategy is proposed to measure the semantic relevant distance for every entity pair in the three entity pair sets and construct knowledge-enhanced entity representations. Afterwards, the entity pairs from each set with top-$k$ semantic inconsistency and consistency are selected to estimate respective knowledge-enhanced entity correlations, by applying negative or positive signed attention with the modal-agnostic features from BSC.
   
   \textbf{Identifying and Linking Entities.} Several public algorithms are utilized to identify textual and visual entities from multimedia posts. Following the work \cite{dun2021kan}, KEC adopts the entity linking solution TAGME to recognize the textual entity set $\boldsymbol{E}_T=\{E_T^i\}_{i=1}^{N_T}$ from the texts. Moreover, KEC exploits the APIs from Baidu OpenAI platform\footnote{https://ai.baidu.com/} to identify the objects, texts and celebrities from the images, which constitute the visual entity set $\boldsymbol{E}_V=\{E_V^i\}_{i=1}^{N_V}$. KEC explores both intra-modality and inter-modality entity correlations by jointly considering the pairwise relationship within and across entity sets $\boldsymbol{E}_T$ and $\boldsymbol{E}_V$. KEC links all possible pairs within the two sets to form two intra-modality entity pair sets $\boldsymbol{E}_{TT}$ and $\boldsymbol{E}_{VV}$ respectively as following:
   \begin{equation}
     \begin{aligned}
       \boldsymbol{E}_{TT} & = \{(E_T^i, E_T^j): 1 \leq i < j \leq N_T\} \\
       \boldsymbol{E}_{VV} & = \{(E_V^i, E_V^j): 1 \leq i < j \leq N_V\}
     \end{aligned}
   \end{equation}
   Similarly, KEC constructs the cross-modality entity pair set $\boldsymbol{E}_{VT}$ as following:
   \begin{align}
     \boldsymbol{E}_{TV} & = \{(E_T^i, E_V^j): 1 \leq i \leq N_T, 1 \leq j \leq N_V\}
   \end{align}
   The semantic relevant distance is measured for each pair in the sets $\boldsymbol{E}_{TT}$, $\boldsymbol{E}_{VV}$ or $\boldsymbol{E}_{TV}$ with the following knowledge relevance reasoning strategy.
   
   \textbf{Semantic Relevant Distance Measurement.} Given an entity pair $(E^u, E^v)$ from arbitary set of $\boldsymbol{E}_{TT}$, $\boldsymbol{E}_{VV}$ or $\boldsymbol{E}_{TV}$, we propose a novel metric $\boldsymbol{D}_s(E^u, E^v)$ to measure the semantic relevant distance of the two entities on a pretrained KG. Different from the metrics in previous works that only considered pairwise feature distance without background contextual knowledge in the KG, the metric $\boldsymbol{D}_s$ is capable to leverage the feature distance in the embedding space as well as the graph distance on the KG topology, which is more appropriate to model the semantic relevant. To calculate $\boldsymbol{D}_s$, KEC firstly find a shortest semantic relevant path $\pi$ in the KG which connects $E^u$ and $E^v$:
   \begin{align}
     \pi :  E^u = E^{w_0} \rightarrow E^{w_1} \rightarrow \ldots \rightarrow E^{w_m} =E^{v}
   \end{align}
   Where $m$ denotes the number of the entities in $\pi$. KEC realizes real-time semantic relevant path searching in a large KG with a modified version of Floyd-Warshall algorithm \cite{hougardy2010floyd} to trade off runtime efficiency, memory consumption and path optimality. \textit{More details of the proposed algorithm can be found in \textbf{Supplementary Materials}.} After obtaining the optimal path $\pi$, the module refines the knowledge-enhanced entity representation $\boldsymbol{h}^u$ for entity $E^u$ as follows:
   \begin{align}
     \boldsymbol{h}^u=\frac{1-\alpha}{1-\alpha^{m+1}}\sum_{i=0}^m \alpha^i \boldsymbol{g}^{w_i}
   \end{align}
   where $\boldsymbol{g}^{w_i}$ is the feature embedding of entity $E^{w_i}$ in the path. $\alpha$ denotes a weight coefficient ($\alpha=0.9$). It's worth noting that $ \boldsymbol{h}^u$ is a path-aware representation, \textit{i.e.}, a different pair $(E^u, E^{v'})$ with path $\pi'$ will yield a different value of $\boldsymbol{h}^u$. Intuitively, $\boldsymbol{h}^u$ absorbs complementary contextual knowledge from all entities along the path $\pi$ by weighted averaging their KG embeddings, with the weight exponentially descending to dilute their contributions as the graph distance increases.  Symmetrically, KEC refines the representation $\boldsymbol{h}^v$ for entity $E^v$ as following:
   \begin{align}
     \boldsymbol{h}^v=\frac{1-\alpha}{1-\alpha^{m+1}}\sum_{i=0}^m \alpha^{m-i} \boldsymbol{g}^{w_i}
   \end{align}
   The semantic relevant distance $\boldsymbol{D}_s(E^u, E^v)$ is calculated as the Euclidean distance between $\boldsymbol{h}^u$ and $\boldsymbol{h}^v$
   \begin{align}
     \boldsymbol{D}_s(E^u, E^v)=\lVert \boldsymbol{h}^u - \boldsymbol{h}^v \rVert_2
   \end{align}
   The concatenated feature $[\boldsymbol{h}^u; \boldsymbol{h}^v]$ is treated as the knowledge-enhanced entity representation for pair $(E^u, E^v)$. The knowledge-enhanced entity representation and semantic relevant distance are further utilized in exploring knowledge-enhanced entity correlation.
   
   
   \textbf{Modeling Knowledge-enhanced Entity Correlation.} KEC models high-order knowledge-enhanced entity correlations in every entity pair set by filtering top-$k$ most relevant/irrelevant pairs and applying positive/negative signed attention with model-agnostic features. Without loss of generality, we elaborate such process for $\boldsymbol{E}_{TV}$ as an example to learn the knowledge-enhanced entity correlation between image and textual modalities.
   
   In order to explore the entity consistency or inconsistency in the pair set $\boldsymbol{E}_{TV}$, KEC constructs two subsets with the top-$k$ most or the least relevant entity pairs in $\boldsymbol{E}_{TV}$. This is achieved by sorting the pairs in $\boldsymbol{E}_{TV}$ with respect to their semantic relevant distances $\boldsymbol{D}_s$.
   The $k$ pairs with smallest distance are chosen as the relevance pair subset $\boldsymbol{E}_{TV,re}$, and the $k$ ones with largest distance as the irrelevance pair subset $\boldsymbol{E}_{TV,ir}$. For better clarification, we denote the semantic relevant distances of entity pairs in $\boldsymbol{E}_{TV,re}$ as $\{d_{re}^i\}_{i=1}^k$ and the corresponding knowledge-enhanced entity representations as $\{\boldsymbol{h}_{re}^i\}_{i=1}^k$. Moreover, we also have $\{d_{ir}^i\}_{i=1}^k$ and $\{\boldsymbol{h}_{ir}^i\}_{i=1}^k$ for the subset $\boldsymbol{E}_{TV,ir}$. The selected knowledge-enhanced entity representations with modal-agnostic semantic features $\hat{\boldsymbol{f}}_T^a$ and $\hat{\boldsymbol{f}}_V^a$ are further fused by utilizing the signed attention mechanism \cite{li2020learning,wu2020hierarchical}, in order to simultaneously capture both high-order consistent and inconsistent entity correlations. Specifically, KEC adopts the positive attention to capture the consistent correlation with respect to the post contents. It takes the concatenated modal-agnostic semantic feature $\hat{\boldsymbol{f}}^a=[\hat{\boldsymbol{f}}_T^a; \hat{\boldsymbol{f}}_V^a]$ (contains modal-shared post content) as the query, and the semantic relevant knowledge-enhanced entity representations $\{\boldsymbol{h}_{re}^i\}_{i=1}^k$ as the key and value to calculate the consistent correlation as following:
   \begin{equation}
     \begin{aligned}
       \{\lambda_{re}^i\}     & = \text{Softmax}\left({\hat{\boldsymbol{f}^a}^T\{\boldsymbol{h}_{re}^i\}}/{\sqrt{2d_e}}\right)                                                \\
       \boldsymbol{f}^{(re)}_{TT} & = \biggl(\sum_{i=1}^k \frac{\lambda_{re}^i}{d_{re}^i} \boldsymbol{h}_{re}^i\biggr)  \bigg/ \biggl({\sum_{i=1}^k \frac{\lambda_{re}^i} {d_{re}^i}}\biggr)
     \end{aligned}
   \end{equation}
   where $2d_e$ is the dimension of $\boldsymbol{h}_{re}^i$. $\{\lambda_{re}^i\}$ denotes the positive attention coefficients. A larger $\lambda_{re}^i$ indicates that the entity pair is more positively semantically associated to the post content. Note that we re-weight the coefficients with $\{d_{re}^i\}_{i=1}^k$ to incorporate semantic relevant distances into the consistency representation $\boldsymbol{f}^{(re)}_{TV}$. A  relevance entity pair with shorter semantic relevant distance influence the learning of consistent correlation more significantly. Simultaneously, KEC utilizes the negative attention to estimate the inconsistency representation $\boldsymbol{f}^{(ir)}_{TV}$ as following:
   \begin{equation}
     \begin{aligned}
       \{\lambda_{ir}^i\}     & = -\text{Softmax}\left(-{\hat{\boldsymbol{f}^a}^T\{\boldsymbol{h}_{ir}^i\}}/{\sqrt{2d_e}}\right)                                              \\
       \boldsymbol{f}^{(ir)}_{TV} & = \biggl(\sum_{i=1}^k \lambda_{ir}^i d_{ir}^i \boldsymbol{h}_{ir}^i\biggr)  \bigg/ \biggl({\sum_{i=1}^k \lambda_{ir}^i d_{ir}^i}\biggr)
     \end{aligned}
   \end{equation}
   The representations $\boldsymbol{f}^{(re)}_{TV}$ and $\boldsymbol{f}^{(ir)}_{TV}$ are then concatenated to form the knowledge-enhanced entity correlation representation $\boldsymbol{f}^{(corr)}_{TV}$ for the entity pair set $\boldsymbol{E}_{TV}$. Similarly, KEC obtains the correlation representation $\boldsymbol{f}^{(corr)}_{TT}, \boldsymbol{f}^{(corr)}_{VV}$ for $\boldsymbol{E}_{TT}$ and $\boldsymbol{E}_{VV}$ with the same mechanism, which are adopted for the final rumor classification.
   
   \subsection{Rumor Classification}
   
   KhiCL leverages basic semantic correlation as well as high-order knowledge-enhanced entity correlation to jointly perform rumor classification. The cross-modal semantic inconsistency and consistency features $\boldsymbol{f}^{(inc)}$ and $\boldsymbol{f}^{(con)}$ and the knowledge-enhanced entity correlation representations $\boldsymbol{f}^{(corr)}_{TT}$, $\boldsymbol{f}^{(corr)}_{TV}$ and $\boldsymbol{f}^{(corr)}_{VV}$ are finally concatenated to form the discriminative  rumor representation, which is further transformed by a FC layer with Sigmoid activation function to predict the rumor probability as following:
   \begin{align}
     \hat{y} = \sigma(\boldsymbol{W}_{c}^T[\boldsymbol{f}^{(inc)}; \boldsymbol{f}^{(con)}; \boldsymbol{f}^{(corr)}_{TT}; \boldsymbol{f}^{(corr)}_{TV}; \boldsymbol{f}^{(corr)}_{VV}]+\boldsymbol{b}_{c})
   \end{align}
   where $\boldsymbol{W}_{c}$ and $\boldsymbol{b}_{c}$ are the parameters of the classifier layer. The overall framework is supervised with a binary cross entropy criterion.
   
   \section{Experiments} 
   \subsection{Experimental Settings}
   
   \begin{table*}[!t]
   \caption{ Performance comparison to the state-of-the-art methods on PHEME and WEIBO datasets.}
     \centering
     \footnotesize
     \resizebox{2.0\columnwidth}{!}{
     \renewcommand{\arraystretch}{1.0}
     \begin{tabular}{c|ccc|cccc|cccc}
       \toprule
       \multirow{2}{*}{\textbf{Method}} & \multicolumn{3}{c|}{\textbf{Modality}} & \multicolumn{4}{c|}{\textbf{PHEME}} & \multicolumn{4}{c}{\textbf{WEIBO}} \\
       \cline{2-12}
       & Text&Image&KG&Acc&Prec&Rec&F1&Acc&Prec&Rec&F1\\
       \midrule
       BERT \cite{devlin2018bert} & \checkmark & & & 0.819 & 0.809 & 0.726 & 0.765 & 0.858 & 0.857 & 0.856 & 0.856\\
       RumorGAN \cite{ma2019detect} & \checkmark & & & 0.783 & 0.785 & 0.783 & 0.782 & 0.867 & 0.866 & 0.866 & 0.866 \\
       EANN \cite{wang2018eann} & \checkmark & \checkmark & & 0.771 & 0.714 & 0.707 & 0.704 & 0.866 & 0.867 & 0.859 & 0.862\\
       MVAE \cite{zhou2020safe} & \checkmark & \checkmark & & 0.776 & 0.735 & 0.723 & 0.728 & 0.824 & 0.828 & 0.822 & 0.823 \\
       SAFE \cite{zhou2020safe} & \checkmark & \checkmark & & 0.807 & 0.787 & 0.789 & 0.791 & 0.851 & 0.849 & 0.849 & 0.849\\
       Singhal \textit{et al.}~\cite{DBLP:conf/www/SinghalPMSK22} &\checkmark&\checkmark&& 0.795 & 0.802 & 0.764& 0.782&0.900&0.882&0.823 &0.847\\ 
       KMGCN \cite{wang2020fake} & \checkmark & \checkmark & \checkmark & 0.812 & 0.775 & 0.753 & 0.764 & 0.861 & 0.864 & 0.856 & 0.860\\
       KAN \cite{dun2021kan} & \checkmark & \checkmark & \checkmark & 0.783 & 0.759 & 0.744 & 0.746 & 0.863 & 0.864 & 0.865 & 0.865 \\
       \textbf{KhiCL} & \checkmark & \checkmark & \checkmark & \textbf{0.874} & \textbf{0.849} & \textbf{0.842} & \textbf{0.846} & \textbf{0.902} & \textbf{0.896} & \textbf{0.894} & \textbf{0.895}\\
       \bottomrule
     \end{tabular}}
   \label{tab:comparation}
   \end{table*}
   
   \textbf{Dataset.} We evaluate the proposed method on two widely-used  benchmark datasets, including PHEME \cite{zubiaga2017exploiting} and WEIBO \cite{jin2017multimodal}. PHEME dataset is collected from the Twitter platform based on five breaking news, each of which contains a set of multimedia posts. WEIBO dataset is collected from XinHua News Agency and the Weibo platform. Each dataset contains a large number of texts and images with labels. Following the work \cite{dun2021kan,qi2021improving,wang2020fake}, these datasets are preprocessed to ensure that each text corresponds to an image. The statistics details of these two datasets are reported in \textit{Supplementary Materials}.  
   
   
   
   \textbf{Implementation Details.} Regarding image content, we employ the pre-trained object detection model \cite{zhu2021tph} and Baidu OpenAI platform APIs with optical character recognition (OCR) and celebrity recognition models to extract the objects, OCR texts and celebrities from the images, which are treated as visual entity mentions and are linked to the corresponding entities in the KG. Regarding text content, the entity linking tool TAGME is applied to link the ambiguous entity mentions in the texts to the corresponding entities in the KG. Freebase \cite{bollacker2008freebase} is introduced as the background KG, where the pre-trained embeddings of the entities with 50 dimension are provided by the method \cite{bordes2013translating}. Moreover, in the textual encoder, we set the length of the input text to at most 128 words, and utilize the pre-trained BERT model to initialize the word embeddings with 768 dimension. In the visual encoder, we use the pre-trained ResNet-50 model on ImageNet \cite{krizhevsky2017imagenet} to extract the visual feature. The dimension of the visual feature $\boldsymbol{f}^i_V$ and textual feature $\boldsymbol{f}^i_T$ are 64. In terms of parameter setting, we set the learning rate of the overall framework to $5\times 10^{-4}$, and fine-tune the BERT model with a learning rate of $1\times 10^{-6}$. The batch size of the input is 16. The number of modality-shared atoms $M$ in the dictionary is $100$. KEC selects top-$3$ most relevant/irrelevant pairs for every entity pair set. Adam optimizer is used to train KhiCL. Accuracy, Precision, Recall and F1 score are employed as the evaluation metrics.

   
   \subsection{Comparison to State-of-the-Art Approaches}
   
   To comprehensively evaluate the proposed method, we compare it with the following state-of-the-art methods on PHEME and WEIBO datasets, including BERT \cite{devlin2018bert}, RumorGAN \cite{ma2019detect}, EANN \cite{wang2018eann}, MVAE \cite{khattar2019mvae}, SAFE \cite{zhou2020safe}, KMGCN \cite{wang2020fake} and KAN \cite{dun2021kan}. In Table~\ref{tab:comparation}, we can observe that KhiCL achieves the best performance of 87.4\% accuracy and 89.6\% accuracy respectively on the two datasets. It improves the second best method SAFE \cite{zhou2020safe} on the PHEME dataset by 6.7\% accuracy, 6.2\% precision, 5.3\% recall and 5.5\% F1 score, and surpasses KAN~\cite{dun2021kan} on the WEIBO dataset by 3.9\% accuracy, 3.2\% precision, 2.9\% recall and 3.0\% F1 score. This comparison demonstrates the effectiveness of the proposed method by jointly modeling the basic semantic correlation and high-order knowledge-enhanced entity correlation. 
   
   
   
   Among these compared methods, BERT \cite{devlin2018bert}, a model that only extracts textual information, which obtains relatively high performance on both datasets, demonstrating its powerful ability to capture the semantics of texts benefited for rumour detection. EANN \cite{wang2018eann} and MVAE \cite{khattar2019mvae} both employ visual and textual multi-modal information, but their performance on PHEME  dataset is inferior to BERT \cite{devlin2018bert}, owing to the fact that they use Text-CNN and Bi-LSTM to learn weaker textual representation, indicating the significant role of text representation in multi-modal rumour detection. SAFE \cite{zhou2020safe} obtains slightly better performance than the EANN and MVAE models, reflecting the benefit of investigating semantic similarity between text and image features. In addition, KMGCN \cite{wang2020fake} and KAN \cite{dun2021kan} outperform most of the compared methods, demonstrating the effectiveness of introducing external complementary knowledge information for multi-modal rumour detection. Compared to all the aforementioned methods, we can attribute the strength of KhiCL to several aspects: 1) The usage of BERT model as a part of the backbone network, resulting in strong textual representation. 2) BSC have superior capacity in learning basic cross-modal semantic inconsistency and consistency between text and image features. 3) The knowledge relevance reasoning strategy within KEC absorbs all complementary contextual knowledge of the relevant entities to learn knowledge-enhanced entity representations. 4) The signed attention mechanism within KEC captures high-order knowledge-enhanced entity correlation.
   
   \subsection{Ablation Studies}
   \textbf{Effectiveness of Components}. To verify the impact of each component within KhiCL, we report the result of ablation study in Table \ref {tab:ablation1} on PHEME dataset. \textit{w/o BSC} refers to KhiCL without using BSC, \textit{w/o KEC} refers to KhiCL without using KEC. The performance of these two KhiCL variants is significantly inferior to the original KhiCL. This indicates that 1) The captured basic cross-modal semantic inconsistency and consistency by BSC play an important role in multi-modal rumor detection, which can effectively promote the discrimination of rumor representation; 2) KEC can effectively model high-order knowledge-enhanced entity inconsistency and consistency, which act as the valuable supplementary  of cross-modal semantic correlation to understand the anomaly in rumors and further improve the detection performance. 
   
   \textbf{Analysis of BSC.} The results in Table \ref{tab:ablation2} show the influence of different components of BSC on PHEME dataset. \textit{w/o FT BERT} refers to BSC extracting textual feature by the pre-trained BERT without fine-tuning, \textit{w/o Align} refers to BSC without using cross-modal joint dictionary to align visual and textual representations (replacing with a shared FC layer). \textit{w/o Se-I} refers to BSC without using the cross-modal semantic inconsistency, \textit{w/o Se-C}: BSC without using the cross-modal semantic consistency. The performance comparison of these BSC variants indicates that 1) Learning more effective textual representation is significant for multi-modal rumor detection; 2) The cross-modal joint dictionary can reduce modality gap and facilitate semantically alignment feature representation; 3) The basic cross-modal semantic consistency provides useful information to judge the authenticity of posts; 4) The basic cross-modal semantic inconsistency also promotes multi-modal information interaction and fusion to improve the detecting performance.   
   
   
   
   
   \begin{table}[!tb]
     \tiny 
      \caption{Evaluation of the effectiveness of each component of KhiCL
   on PHEME dataset.}
      \begin{center}
   \resizebox{\columnwidth}{!}{
        \renewcommand{\arraystretch}{1.0}
         \begin{tabular}{c|cccc}
   
            \hline
         \textbf{Model}&\textbf{Acc}&\textbf{Prec}&\textbf{Recall}&\textbf{F1}\\
            \hline
            \multirow{5}{*}{\textbf{\begin{sideways}\end{sideways}}} 
             \textbf{KhiCL} & \textbf{0.874} & \textbf{0.849} & \textbf{0.842} & \textbf{0.846} \\
                w/o BSC & 0.755 & 0.700 & 0.687 & 0.694 \\
             w/o KEC & 0.824 & 0.787 & 0.789 & 0.788 \\

   \hline
      
         \end{tabular}}
         \label{tab:ablation1}
      \end{center}
      \vspace{-.3cm}
   \end{table}
   
   \begin{table}[!t]
   \scriptsize 
      \caption{Evaluation of the influence of different components of BSC on PHEME dataset.}
      \begin{center}
      \resizebox{\columnwidth}{!}{
        \renewcommand{\arraystretch}{1.0}
         \begin{tabular}{c|cccc}
            \hline		\textbf{Model}&\textbf{Acc}&\textbf{Prec}&\textbf{Recall}&\textbf{F1}\\
            \hline
            \multirow{5}{*}{\textbf{\begin{sideways}\end{sideways}}} 
                \textbf{BSC} & \textbf{0.874} & \textbf{0.849} & \textbf{0.842} & \textbf{0.846} \\
                w/o FT BERT & 0.863 & 0.840 & 0.818 & 0.829 \\
                w/o Align & 0.849 & 0.819 & 0.811 & 0.815 \\
             w/o Se-I & 0.854 & 0.822 & 0.824 & 0.833 \\
                w/o Se-C & 0.820 & 0.781 & 0.784 & 0.782 \\
   
   \hline
      
         \end{tabular}}
         \label{tab:ablation2}
      \end{center}
      \vspace{-.3cm}
   \end{table}
   
   \begin{table}[!t]
      \caption{Evaluation of the influence of different components of KEC on PHEME dataset.}
      \begin{center}
       \tiny 
      \resizebox{\columnwidth}{!}{
        \renewcommand{\arraystretch}{1.0}
         \begin{tabular}{c|cccc}
            \hline		
            \textbf{Model}&\textbf{Acc}&\textbf{Prec}&\textbf{Recall}&\textbf{F1}\\
            \hline
            \multirow{5}{*}{\textbf{\begin{sideways}\end{sideways}}}  
               \textbf{KEC} & \textbf{0.874} & \textbf{0.849} & \textbf{0.842} & \textbf{0.846} \\
                w/o Path & 0.854 & 0.821 & 0.830 & 0.825 \\
                w/o E-I & 0.833 & 0.798 & 0.798 & 0.798 \\
                w/o E-C & 0.856 & 0.848 & 0.789 & 0.817 \\
   \hline
      
         \end{tabular}}
         \label{tab:ablation3}
      \end{center}
      \vspace{-.3cm}
   \end{table}
   
   
   
   \textbf{Analysis of KEC.} We conduct the ablation experiments to analyze the influence of different components of KEC on PHEME dataset in Table \ref{tab:ablation3}.  \textit{w/o Path} denotes KEC without using the knowledge relevance reasoning strategy (replacing with direct pairwise feature distance), $\textit{w/o E-I}$ denotes KEC without utilizing the knowledge-enhanced entity inconsistency, $\textit{w/o E-C}$ denotes KEC without utilizing the knowledge-enhanced entity consistency. From the performance comparison, we can observe that 1) The knowledge relevance reasoning strategy can find the optimal semantic relevant path, and absorb the background complementary knowledge in the KG to preferably infer the entity correlation. 2) The positive and negative signed attention mechanism can effectively capture and fuse knowledge-enhanced consistency and inconsistency information, each of which has the powerful influence in detection multi-modal rumor. 3) Compared to the entity consistency, entity inconsistency provide more crucial information to judge the authenticity of a post.
   
\begin{figure}[!tbp]
      \centering
      \includegraphics[width=\linewidth]{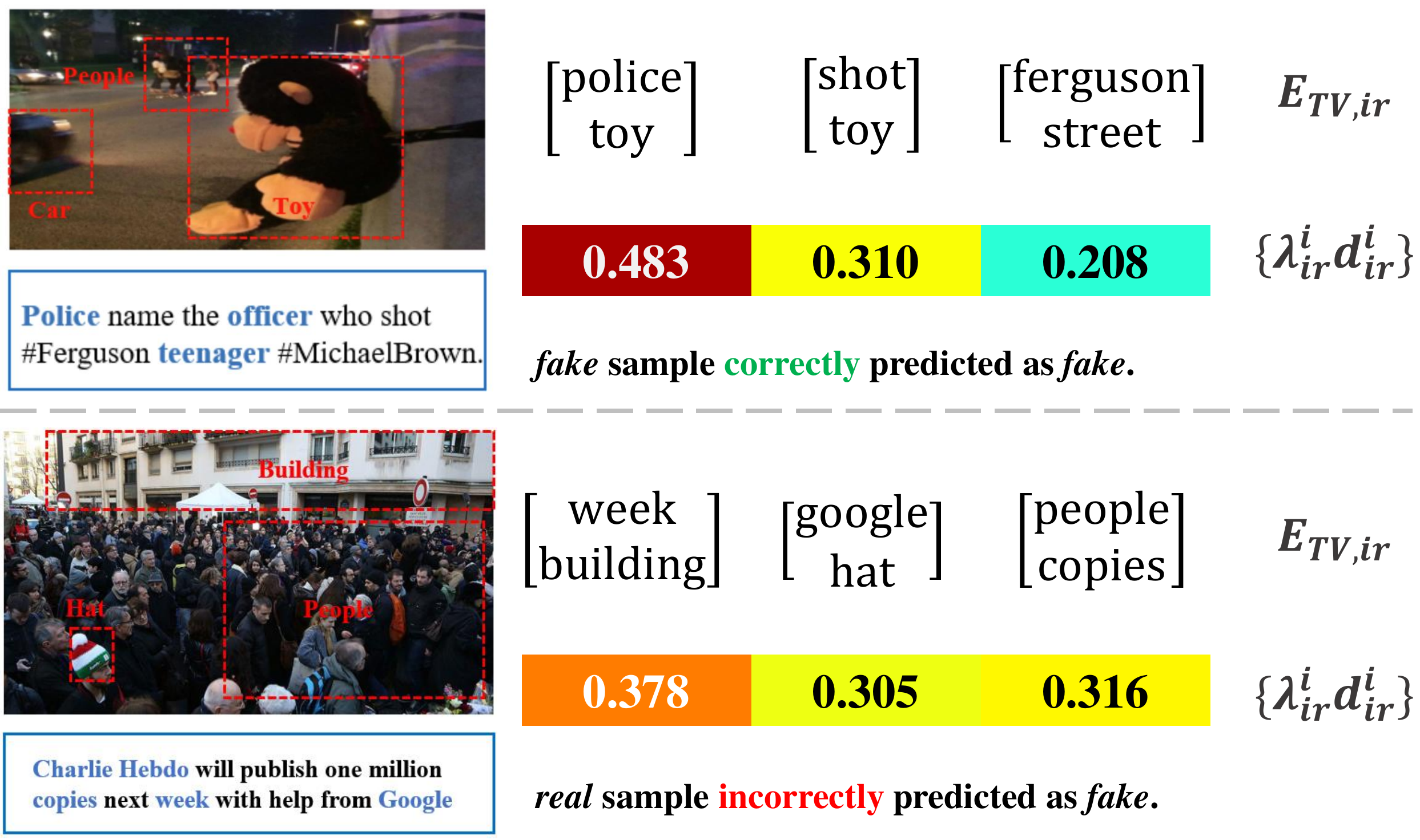}
      \caption{Visualization results of a successfully predicted and an incorrectly classified rumor posts from the PHEME dataset.}
      \label{fig:entity}
   \end{figure}

   \textbf{Visualization Results.} Figure~\ref{fig:entity} visualizes the textual-visual irrelevant entity pair set $\boldsymbol{E}_{TV,ir}$ and their associated negative attention values $\{\lambda_{ir}^id_{ir}^i\}$ of two posts. In the first row, KhiCL effectively captures the unusual pair ``police-toy'' and emphasizes its attention weight, leading to the correct result of predicting as fake, which demonstrates the success of KhiCL in modeling high-order correlation. However, in the second row KhiCL extracts visual entities of low quality and is further misled to make incorrect judgement. This reveals the defect of entity extractor could be one potential
   source of classification failure.
   
   

   \section{Conclusion}
   In this work, we propose a novel knowledge-enhanced hierarchical information correlation learning approach (KhiCL) for multi-modal rumor detection by jointly modeling the basic semantic correlation and high-order knowledge-enhanced entity correlation. Concretely, we propose a Basic Semantic Correlation Module. It introduces a cross-modal joint dictionary to reduce the large modality gap and learn the aligned visual and textual representations, and further utilizes a cross-modal fusion layer to estimate basic cross-modal semantic correlation. Moreover, we propose a Knowledge-Enhanced Entity Correlation Module to learn knowledge-enhanced entity representations via knowledge relevance reasoning strategy, and model the knowledge-enhanced entity correlation of intra-modality and inter-modality entity pairs by signed attention mechanism, serving as the significance supplement of the cross-modal semantic correlation at the feature level. Extensive experiments on two  benchmarks \emph{i.e.}, PHEME  and WEIBO datasets validate the effectiveness of KhiCL.

   {\small
   \bibliographystyle{ieee_fullname}
   \bibliography{main}
   }

\end{document}